\pdfoutput=1

\documentclass[11pt]{article}

\usepackage{EACL2023}

\usepackage{times}
\usepackage{latexsym}

\usepackage[T1]{fontenc}

\usepackage[utf8]{inputenc}

\usepackage{microtype}

\usepackage{inconsolata}

\usepackage{graphicx} 
\usepackage{multirow} 
\newcommand{\sm}[1]{{\color{black} #1}}
\usepackage{enumitem} 
\title{Frustratingly Easy Sentiment Analysis of Text Streams:\\
Generating High-Quality Emotion Arcs Using Emotion Lexicons}

\author{Daniela Teodorescu \\
  University of Alberta\\
  \texttt{dteodore@ualberta.ca} \\\And
  Saif M. Mohammad\\
National Research Council Canada\\
  \texttt{saif.mohammad@nrc-cnrc.gc.ca} \\}
 
\begin{document}
\maketitle
\begin{abstract}
\sm{Automatically generated emotion arcs---that capture how an individual or a population feels 
over time---are widely used in industry and  
research. 
However, there is little work on evaluating the generated arcs. This is in part due to the difficulty of establishing the true (gold) emotion arc. Our work, for the first time, systematically and quantitatively evaluates automatically generated emotion arcs. We also compare two common ways of generating emotion arcs:
Machine-Learning (ML) models 
and Lexicon-Only (LexO) methods.}
Using a number of diverse datasets, we systematically study the relationship between the \textit{quality of an emotion lexicon} and the quality of the emotion arc that can be generated with it. We also study the relationship between the \textit{quality of an instance-level emotion detection system} (say from an ML model) and the quality of emotion arcs that can be generated with it. We show that despite being markedly poor at instance level, LexO methods 
are highly accurate at generating emotion arcs by aggregating information from hundreds of instances. This has wide-spread implications for commercial development, as well as research in psychology, public health, digital humanities, etc.\@ that values simple interpretable methods and disprefers the need for domain-specific training data, programming expertise, and high-carbon-footprint models. 
\end{abstract}

\section{Introduction}
\setitemize[0]{leftmargin=5mm}
\setenumerate[0]{leftmargin=*}

Commercial applications as well as research projects often benefit from accurately tracking the emotions associated with an entity over time \cite{cellphone,policy,socialmedia,Seidl2020ThePO,somasundaran-etal-2020-emotion,kim-etal-2017-investigating}. Here, the goal is not to determine some absolute value of emotion associated with the entity, but rather whether the degree of that emotion has remained steady, increased, or decreased from one time step to the next. For example, how has the anger towards the state of public infrastructure in a city changed from one month to another over the last ten years. In this example, the time steps of consideration are months, but other time steps commonly used are days, weeks, and years. 
This series of time step–emotion value pairs, which can be represented as a time-series graph, is often referred to as an \textit{emotion arc} \cite{mohammad-2011-upon,emotionarcs}.

When the amount of data involved is large enough that human annotations of emotions are prohibitive, then one can employ automatic methods to estimate emotion arcs. Automatic methods will be less accurate than human assessments, but 
can be applied at scale and easily adjusted to changes in needs (e.g., tracking new, different, or additional entities of interest). 

\noindent The input to these systems are usually:\\[-22pt]
\begin{itemize}
\item The text of interest where the individual sentences (or instances) are temporally ordered; possibly through available timestamps indicating when the instances were posted/uttered: e.g., all the tweets relevant to (or mentioning) a government policy along with their meta-data that includes the date and time of posting.\\[-20pt]
\item The emotion dimension/category of interest: e.g., anxiety, anger, valence, arousal, etc. \\[-20pt]
\item The time step granularity of interest (e.g., days, weeks, months etc.)\\[-20pt]
\end{itemize}
The automatic methods usually employ the steps listed below to determine the time step--emotion value pairs (the emotion arc):\\[-20pt]
\begin{itemize}
\item Suitable pre-processing of the text (e.g., converting text to lowercase, removing urls and numbers, tokenizing text, etc.).\\[-20pt]
\item Apply emotion labels to units of text. 
Two common approaches for labeling units of text are: (1) \textit{The Lexicon-Only (LexO) Method:} to label words using emotion lexicons and (2) \textit{The Machine-Learning (ML) Method:} to label whole sentences using supervised ML models (with or without using emotion lexicons).\\[-20pt]
\item Aggregate the information to compute time step–emotion value scores; e.g.,
if using the lexicon approach: for each time step,
compute the percentage of angry words or average valence of the words in the target text pertaining to each time step, 
and if using the ML approach: for each time step, compute
the percentage of angry sentences or average valence of the sentences  
for each time step. \\[-20pt]
\end{itemize}
\noindent The time steps used can be non-overlapping, e.g., months of a year, but they can also be overlapping, e.g., ten-day time steps starting at every day of a year. Here, every adjacent time step has nine overlapping days. Overlapping steps produce smoother emotion arcs, and thus are preferred in some applications. \sm{The number of textual instances (usually sentences or tweets) pertaining to a time step are referred to as the \textit{size of the time step} or {\it bin size}.} 
In case the data does not come with associated time stamps, but simply a temporal order from beginning to end (such as the text in a novel), then often a time step is determined by a pre-chosen fixed amount of textual units (e.g., 200 words, 100 sentences, or one chapter). 
Thus, here all bins have the same size.

\sm{Despite their wide-spread use in industry and research, there is little work on evaluating the generated emotion arcs. This is in part due to the difficulty of establishing the true (gold) emotion arc.} We also know little about how best to aggregate information when using emotion lexicons; e.g., is it better to use coarse or fine-grained lexicons, should we ignore slightly emotional words, and how to handle out of vocabulary (OOV) terms (terms not in the emotion lexicon). 
\sm{Our work, for the first time, systematically and quantitatively evaluates automatically generated emotion arcs. We also compare the ML and LexO methods for generating arcs.}
\sm{We conduct experiments to} study the relationship between the \textit{quality of an emotion lexicon} and the quality of emotion arcs that can be generated with it, by varying various parameters involved. We also study the relationship between the \textit{quality of an instance-level emotion detection system} (say, from an ML model) and the quality of emotion arcs that can be generated with it.
Specifically, we explore these research questions:\\[-20pt]
\begin{enumerate}
    \item How accurate are the ML and LexO methods for emotion arc generation? \\[-20pt]
    \item Are the gains obtained using an ML model  
    enough to offset its accessibility, interpretability, financial, and environmental costs?\\[-20pt]
    \item How to best use an emotion lexicon to generate emotion arcs?\\[-20pt]
    \item How does the quality of the generated emotion arcs change with the level of aggregation (number of instances binned 
    together 
    to form a time step)? These results will help one judge how to balance emotion arc quality and the level of detail/granularity at which emotion arcs should be generated. \\[-20pt]
    \item How good are existing emotion lexicons  
    in terms of generating corresponding emotion arcs? Is it harder to generate accurate arcs for some emotions compared to others? These results will shed light on the nature of different emotions.\\[-20pt]
    \item How does the granularity of the association scores available in an emotion lexicon impact the quality of emotion arcs?
    Does the use of fine-grained lexicons 
    containing  
    real-valued emotion association scores lead to substantially better emotion arcs than with the use of coarse grained lexicons that only indicate whether a word is associated with an emotion or not associated. These results will help one judge whether to invest in the costlier fine-grained lexicons for a language.\\[-20pt]
    \item For a given emotion dimension or category, should we use the full set of associated words or only those words that have an association greater than some threshold? Does this threshold vary widely across emotions or is it roughly within a narrow band of association scores?\\[-20pt]
    \item How best to handle out of vocabulary terms (words not in an emotion lexicon)?\\[-20pt]
    \item Is it viable to use automatic translations of emotion lexicons to generate emotion arcs in a low-resource language?\\[-20pt]
\end{enumerate}
\noindent We conduct these experiments on a diverse set of datasets, including: tweets, SMS messages, personal diary posts, and movie reviews. We make the data and code freely available. \sm{Notably, the code allows users to easily generate high-quality emotion arcs for their text of interest.}\footnote{\sm{\url{https://github.com/dteodore/EmotionArcs}}}

\section{Related Work}
Several systems for generating emotion arcs have been previously proposed in various works, most commonly for creating emotion arcs for characters in story dialogues. 
\citet{mohammad-2011-upon} analyzed the flow of emotions across various novels and books using emotion lexicons. \citet{kim-etal-2017-investigating} built on this work by creating emotion arcs to determine emotion information for various genres using the NRC Emotion Lexicon. 
\citet{somasundaran-etal-2020-emotion} generated emotion arcs to analyze, and assess the quality of narratives written by students using an event polarity lexicon \cite{10.5555/3504035.3504742} and a connotation lexicon \cite{feng-etal-2013-connotation} for extracted events. 
\citet{moviedialogues} recently analyzed emotion arcs for characters in movie dialogues using emotion lexicons. 
\citet{Bhyravajjula2022MARCUSAE} create a pipeline for plotting a character's arc using ML methods such as a fine-tuned RoBERTa \cite{liu2020roberta} model. \citet{brahman-chaturvedi-2020-modeling} model emotion arcs for protagonists in generated stories using supervised and reinforcement learning methods.
Further, other work has approached constructing emotion arcs using reinforcement learning, as \citet{brahman-chaturvedi-2020-modeling} do for protagonists in stories. Lastly 
emotion arcs 
have been employed 
to assist
writers with creating stories which match an emotion arc that readers are expecting and desire \cite{writenovels}. However, there is surprisingly little work on arc evaluation.
A key reason for this is that it is hard to determine the true emotion arc of a story from data annotation. It is hard for people to read a large amount of text, say from a novel, and produce an emotion arc for it.

Emotion arcs commonly employed in commerce and social media research are fundamentally different from the arcs in novels. There, one is often interested in the arcs associated with posts that mention a pre-chosen product, such as a certain brand of cellphone \cite{cellphone}, government policy \cite{policy}, a person \cite{socialmedia}, or entity such as Uber \cite{Seidl2020ThePO}. For example, an emotion arc of tweets that mention the latest iPhone. Here, the different instances are not part of a coherent narrative. Thus, while it would be difficult to find a literary theorist who would agree that the true emotion arc of a novel is simply the average of the emotions of the constituent sentences, it is arguably more persuasive
to consider the gold emotion arcs to be simply the average of the emotion scores of all the tweets pertaining to the time steps. 
For example, the average human-annotated emotion scores of the iPhone tweets posted every day.

\begin{table*}[h]
\centering
{\small
\begin{tabular}
{llllll}
\hline
\multicolumn{1}{l}{\textbf{Dataset}} &
\multicolumn{1}{l}{\textbf{Source}} &
\multicolumn{1}{l}{\textbf{Domain}} &
\multicolumn{1}{l}{\textbf{Dimension}} &
\multicolumn{1}{l}{\textbf{Label Type}} & \multicolumn{1}{l}{\textbf{\# Instances}}\\ \hline
Movie Reviews Categorical& \citet{socher-etal-2013-recursive} & movie & valence & categorical (0, 1) & 11272\\
Movie Reviews Continuous     & \citet{socher-etal-2013-recursive}  & reviews  & valence   & continuous (0 to 1) & 11272\\[3pt]

SemEval 2014  
   & \citet{rosenthal-etal-2014-semeval} & Multiple$^*$ & valence & categorical (-1, 0, 1) & Multiple$^*$ \\[3pt]

SemEval 2018 (EI-OC) & \citet{SemEval2018Task1} & tweets & anger, fear,  & categorical (0, 1, 2, 3) & 3092, 3627, \\
     &  &   & joy, sadness   &  & 3011, 2095 \\[3pt]

SemEval 2018 (EI-Reg) & \citet{SemEval2018Task1} & tweets & anger, fear & continuous (0 to 1) & 3092, 3627,\\
     &  &   & joy, sadness   &  &  3011, 2095 \\[3pt]

SemEval 2018 (V-OC) &  \citet{SemEval2018Task1} & tweets & valence & categorical (-3,-2,...3) & 2567\\
SemEval 2018 (V-Reg) &  \citet{SemEval2018Task1} & tweets & valence & continuous (0 to 1) & 2567\\\hline

\end{tabular}
}
\vspace*{-1mm}
\caption{Dataset descriptive statistics. The No. of instances includes the train, dev, and test sets for the Sem-Eval 2014 Task 9 and the Sem-Eval 2018 Task 1 (EI-OC, EI-Reg, V-OC and V-Reg). Multiple$^*$: The SemEval 2014 dataset has collections of LiveJournal posts (1141), SMS messages (2082), regular tweets (15302), and sarcastic tweets (86) for a total of 18611 instances. The Movie Reviews Categorical dataset is often also referred to as SST-2.} 
 \label{tab:datasets}
\end{table*}

\section{Experimental Setup to Evaluate Automatically Generated Emotion Arcs}
We 
construct \textit{gold emotion arcs} from existing datasets where individual instances are manually annotated for emotion labels. Here, an instance could be a tweet, a sentence from a customer review, a sentence from a personal blog, etc. \sm{Depending on the dataset, manual annotations for an instance may represent the emotion of a speaker, 
or 
sentiment
towards a product or entity.}
For a pre-chosen  
bin size 
say 100 instances per bin, we compute the \sm{gold} emotion score by taking the average of the human-labeled emotion scores of the instances in that bin (in-line with the commerce and social media use cases discussed earlier). We move the window forward by one instance, compute the average 
in that bin, and so on.\footnote{\sm{Using larger window sizes does not impact  conclusions.}} 

\sm{We standardized all of the emotion arcs (aka z-score normalization) so that the gold emotion arcs are comparable to automatically generated arcs.\footnote{\sm{Subtract the mean from the score and divide by the standard deviation (arcs then have zero mean and unit variance).}}}
Finally, we evaluate an automatically generated emotion arc by computing the linear correlation between the system-predicted emotion arc and the gold arc using 
\sm{Spearman correlation \cite{spearman1987proof}.
High correlation implies greater fidelity: no matter what the overall shape of the gold emotion arc, when it goes up (or down), the predicted emotion arc also
goes up (or down).}\footnote{\sm{Shape of the emotion arc: Different emotion arc shapes
can be generated by ordering the data differently. Whether using the shapes using the time-stamps
associated with the instances or any other method, we found the same Spearman correlations (and conclusions)
as described in the paper ahead.}}

\begin{table*}[ht]
\centering
{\small
\begin{tabular}
{llllr}
\hline
\multicolumn{1}{l}{\textbf{Lexicon}} &
\multicolumn{1}{l}{\textbf{Source}} &
\multicolumn{1}{l}{\textbf{Categories / Dimensions}} &
\multicolumn{1}{l}{\textbf{Label Type}} &
\multicolumn{1}{l}{\textbf{\# Terms}}\\ \hline

NRC EmoLex   & \citet{Mohammad13} & \textbf{anger, joy, fear, sadness,} 4 others  & categorical (0, 1) & 14,154\\

NRC EIL  & \citet{LREC18-AIL, mohammad2020practical} & \textbf{anger, joy, fear, sadness,} 4 others & continuous  (0 to 1) & 14,154\\[3pt]

 NRC VAD  & \citet{vad-acl2018} & \textbf{valence}, arousal, dominance & continuous (-1 to 1) & 20,007\\
  & & \textbf{valence}, arousal, dominance & categorical (-1, 0, 1) & 20,007\\
\hline
 
\end{tabular}
}
\vspace*{-1mm}
\caption{Lexicons used this study. 
The subset of emotions explored in our experiments are marked in bold.}
\vspace*{-3mm}
\label{tab:lexicons}
\end{table*}

Table \ref{tab:datasets} shows details of the instance-labeled \sm{English (Eng)} test data used in our experiments. Observe that the datasets are of two kinds: those with categorical labels such as the SemEval 2014, which has -1 (negative), 0 (neutral), and 1 (positive), as well as those with continuous labels such as SemEval 2018 (EI-Reg), which has real-valued emotion intensity scores (for four emotions) between 0 (lowest/no intensity) and 1 (highest intensity). 

\noindent We use the above evaluation setup to study:\\[-20pt]
\begin{enumerate}
\item The relationship between the quality of \textit{word-level} emotion labeling
(the usual lexicon approach) on the generated emotion arcs  (\S 4).\\[-20pt]
\item  The relationship between the quality of \textit{instance-level} emotion labeling
(the usual ML approach) on the generated emotion arcs (\S 5). 
\end{enumerate}
\noindent Table \ref{tab:lexicons} shows the lexicons we used to generate emotion arcs studied in \S 4. These lexicons are widely used to support sentiment analysis and have both categorical and real-valued versions, which is useful to study whether using fine-grained lexicons leads to markedly better emotion arcs than when using coarse categorical lexicons (keeping the vocabulary of the lexicon constant). 
We simulate performance of ML models with varying accuracies using an Oracle system described in \S 5.

\sm{Finally, to determine whether using automatic translations of English lexicons into relatively less-resource languages is a viable option, we experiments with the Arabic (Ar) and Spanish (Es) emotion-labeled tweet data from SemEval 2018 Task 1 \cite{SemEval2018Task1}. Similarly to the English dataset, these contain original tweets in the language (not translated) with emotion labels by native speakers, on categorical and continuous scales. Details of these datasets are in Tables \ref{tab:datasets_arabic} and \ref{tab:datasets_spanish} in the Appendix.}

\sm{In all, we conducted experiments on 36
emotion-labeled datasets from three languages, with labels that are categorical and continuous for five affect categories.
This helps determine whether the inferences we draw are limited to individual datasets or more general. 
While the core 
experiments are presented in the main paper,  
other result tables are available in the Appendix. The results on individual datasets also establish key benchmarks; useful to 
practitioners for estimating the quality of the arcs under various settings.}

\section{LexO Arcs: Emotion Arcs Generated from Counting Emotion Words}
As discussed in the Introduction, the quality of an emotion lexicon is expected to impact the quality of the automatically generated emotion arcs. 
We now describe experiments  
where we systematically vary various parameters associated with using an emotion lexicon to generate emotion arcs.

We begin  
by generating arcs of valence (or sentiment) 
 using lexicons of different score granularities (categorical labels and continuous labels), two different methods of handling out of vocabulary (OOV) terms,
 and using various levels of aggregation or bin sizes (groupings of instances used to generate the emotion scores for an arc)
which we describe in
 Section \ref{sec:valence}.
For our experiments we explored bin sizes of 1, 10, 50, 100, 200, and 300. 
The two OOV handling methods explored were: 1. Assign label NA (no score available) and disregard these words, and 2. Assign 0 score (neutral or not associated with emotion category), thereby, leading to a lower average score for the instance than if the word was disregarded completely. 

We then evaluate how closely the arcs correspond to the gold valence arcs.
The same experiment is then repeated for various other emotion categories, to see if we find consistent patterns (Section \ref{sec:emotions}). 
For each emotion category or dimension, we generate the predicted emotion arcs using the corresponding existing
emotion lexicons listed in Table \ref{tab:lexicons}. 
In Section \ref{sec:thresh}, we show the impact of using emotion lexicons more selectively. Specifically,
by only using entries that have an emotion association greater than some pre-determined threshold.

\begin{figure*}[t!]
    \centering
     \includegraphics[width=\textwidth]{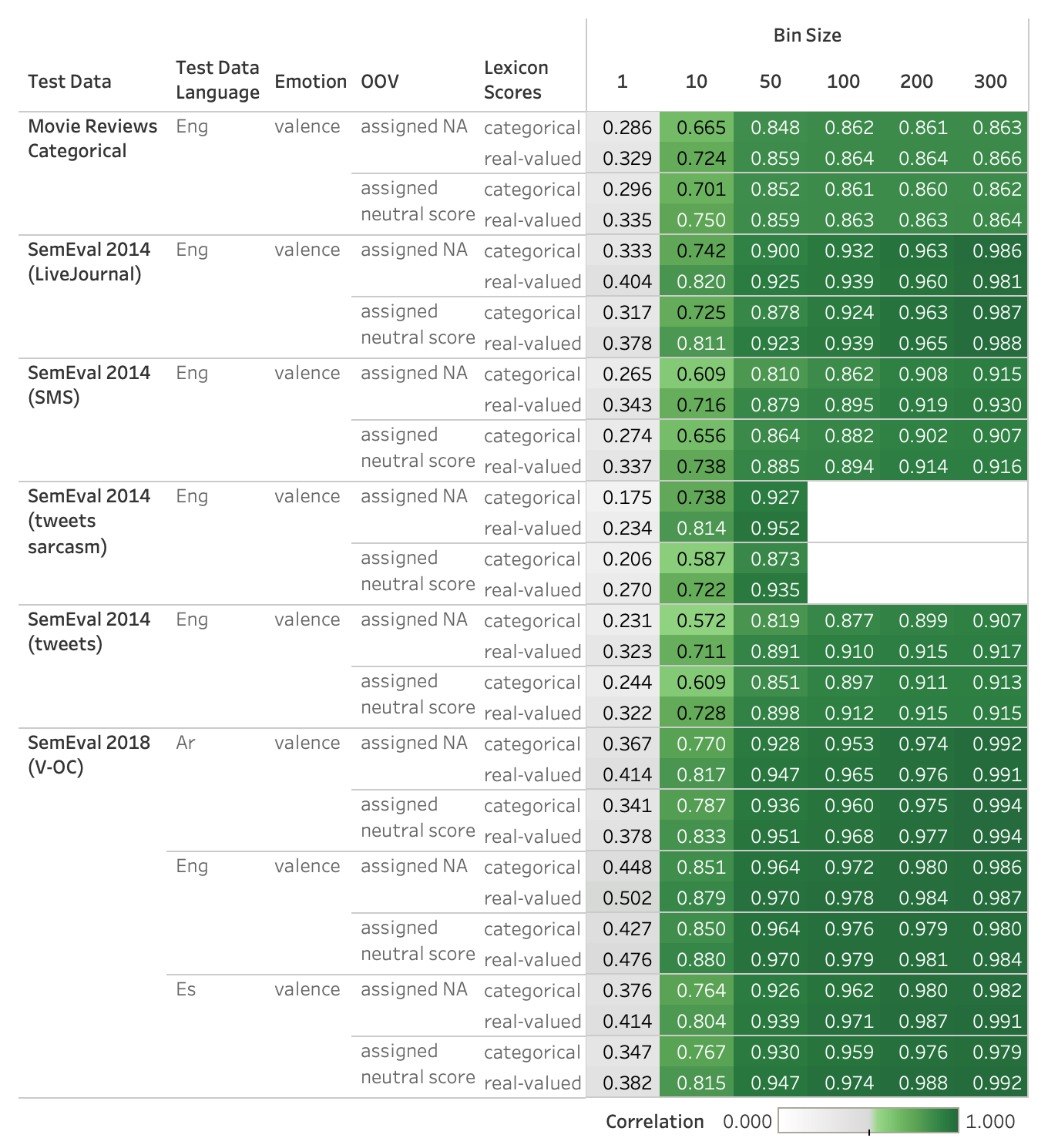}
    \caption{Valence:
    \sm{Spearman} correlations between 
    arcs generated using lexicons and 
    gold arcs
    created from \textit{categorically} labeled  test data.}
    \label{fig:v_cat_results}
    \vspace*{-4mm}
\end{figure*}

\subsection{Valence: Impact of Bin Size, Lexicon Granularity, OOV Handling}
\label{sec:valence}

Figure \ref{fig:v_cat_results} shows the correlations between valence arcs generated using information from the lexicons and the gold valence arcs  
created from the categorically labeled test data. 
Figure \ref{fig:v_reg_results} shows the results for the same experiments but 
for gold arcs from
continuously labeled valence test data.

\noindent \textbf{Bin Size:}
 Overall, across datasets and regardless of the type of lexicon used and how OOV words are handled, 
 increasing the bin size dramatically improves correlation with the gold arcs. In fact, with bin sizes as small as 50, many of the generated arcs have correlations above 0.9. With bin size 100 correlations are around 0.95, and approach 
 \sm{high 0.90's to 0.99}
 with bin sizes of 200 and 300.
 
 \noindent {\it Discussion:} For many social media applications, one has access to tens of thousands, if not millions of tweets. There, it is not uncommon to have time-steps (bins) that include thousands of tweets. Thus it is remarkable that even with relatively small bin sizes of a few hundred tweets, the simple lexicon approach is able to obtain very high correlations. Of course, the point is not that the lexicon approach is somehow special, but rather that aggregation of information can very quickly generate high quality arcs, even if the constituent individual emotion signals are somewhat weak.

\sm{ \noindent  \textbf{Translated Lexicons:}
Using Arabic and Spanish translations of the English emotion lexicons 
also resulted in high correlation scores for larger bin sizes (just as in the case of the English datasets).

\noindent \textit{Discussion:} Thus, one key outcome of these experiments is that for low-resource languages,
where labeled training data and emotion lexicons are scarce, using translations of English lexicons
to generate emotion arcs is particularly attractive. This approach can thus unlock the potential
for affect-related psychology, health science, and digital humanities research for a vast number of languages.} 

\noindent  \textbf{Categorical vs. Real-Valued Lexicons:}
Using a real-valued lexicon obtains higher correlations across bin sizes, methods for processing OOV terms, and datasets. 
The difference is marked for very small bin sizes (such as 1 and 10) but progressively smaller for higher bin sizes. 

\noindent {\it Discussion:} Entries from real-valued lexicons carry more fine-grained emotion information, and it is likely that this extra information is especially helpful when emotion scores are determined from very little text (as in the case of small bins), but less useful for larger bin sizes where even coarse information from the additional text is sufficient to obtain high correlations. 

\noindent \textbf{Processing OOV Terms:}
Each of the two methods for processing OOV words performed slightly better than the other method, roughly an equal number of times. 
Overall, for larger bin sizes such as 200 and 300 both methods obtained similar (very high) correlations.

\noindent {\it Discussion:} 
Results for OOV handling are some what mixed, and might merit further enquiry in sparse data scenarios, but the issue is not of practical relevance for
bin sizes of 100 and above, 
where either approach leads to similar results.

\subsection{Anger, Fear, Joy, Sadness: Impact of Bin Size, Lexicon Granularity, OOV Handling}
\label{sec:emotions}

\begin{figure*}[t]
    \centering
    \includegraphics[width=0.95\textwidth]{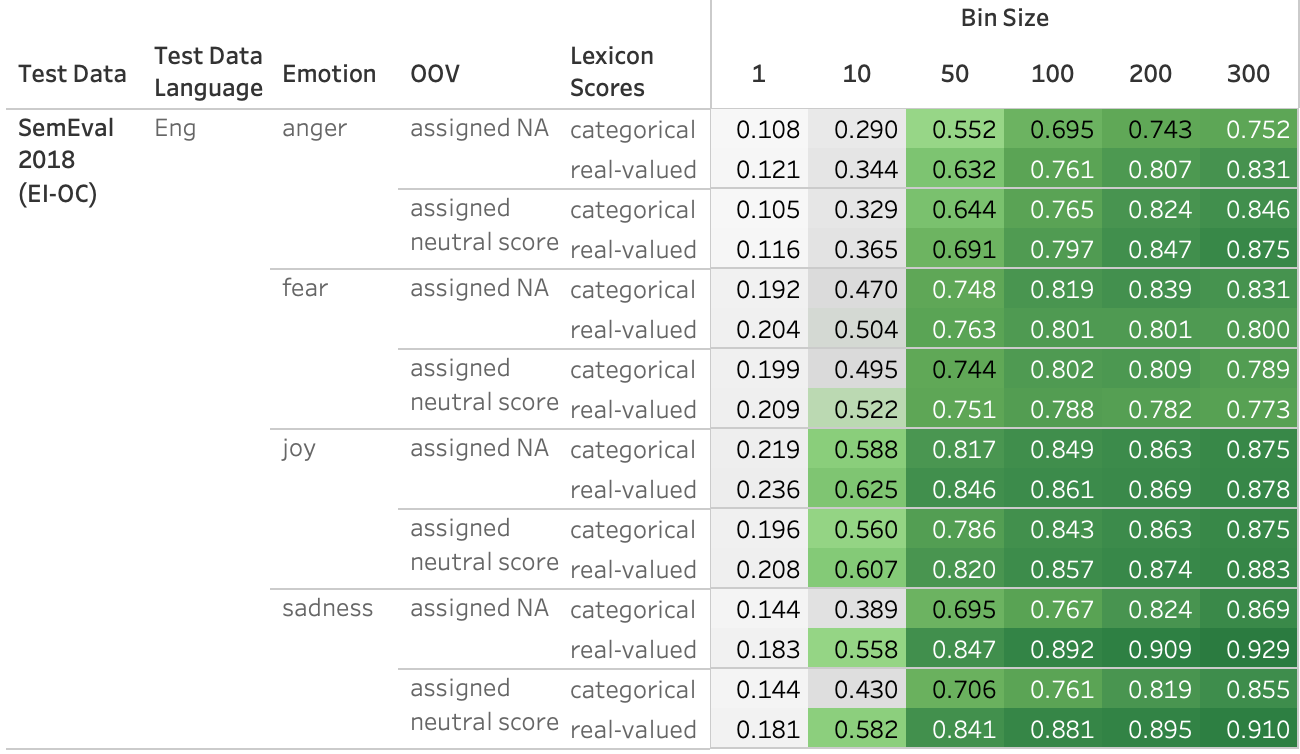}
    \caption{Anger, fear, joy and sadness: 
    \sm{Spearman}
    correlations between arcs generated using lexicons and 
    gold arcs
    created from \textit{categorically} labeled test data.}
   \vspace*{-3mm}
    \label{fig:e_cat_results}
 
\end{figure*}

Figure \ref{fig:e_cat_results} shows the correlations between emotion arcs generated using information from the lexicons and the gold emotion arcs  
created from the categorically labeled emotion test data (for anger, fear, joy, and sadness).
\sm{We observed similar results on arcs generated from the continuously labeled emotion test data as observed on the gold arcs created from categorical data. We include those results  in the Appendix (Figure \ref{fig:e_reg_results}).}

\noindent \textbf{Bin Size:}
We observe the same trend for anger, fear, joy, and sadness, as we did for valence earlier: with increased bin sizes correlations with arcs generated from humans annotations increases substantially. However, we note that overall the correlations for these emotion categories are lower than for valence, with values ranging from
\sm{0.10 and 0.20 for Ar, mid 0.20's for Eng, and 0.30 for Es}
at bin size 1 
\sm{to mid 0.80--high 0.90's for bin sizes 200 and 300.}

\noindent {\it Discussion:} Past work has shown that 
emotion detection at the instance level is less accurate for emotions such as anger and sadness compared to valence \cite{SemEval2018Task1}. Work in psychology has also shown valence to be the prime dimension of connotative meaning \cite{Osgood57,russell2003core,russell1977evidence}. 
Thus, the lower correlations we observe for categorical emotions \sm{(especially for Ar and Eng)} compared to valence aligns with those findings; suggesting that there are more overt, easier to discern, clues for valence. 

\noindent \textbf{Categorical vs. Real-Valued Lexicons:}
We observe the same trend for the anger, fear, joy and sadness, as we did for valence: Generally, the real-valued lexicons obtain higher correlations than the categorical lexicons (this is especially prominent for \sm{Eng} sadness and \sm{Es fear}). 
(A notable exception is \sm{anger Ar for bin size 1, 50, 100 and} fear \sm{Eng} for bin size \sm{100} 
and above, and lastly sadness Es for bin size 50 and above based on the lexicon scores type.)
Also, unlike as in the case of valence 
\sm{and for noted exceptions,}
real-valued lexicons are markedly better for higher bin sizes as well.

\noindent {\it Discussion:} Greater benefit from the real-valued lexicons in case of anger, sadness, etc.\@, as compared to valence, is probably because they are relatively harder to identify; and so the extra information from real-valued lexicons is especially beneficial. 

\noindent \textbf{Processing OOV Terms:}
Figure \ref{fig:e_cat_results} shows that depending on the emotion, a certain method for handling OOV words produces higher correlation scores. For example, the assigned neutral score obtains higher correlations for \sm{Eng} anger with bin sizes 10 and up,
whereas assigning NA to OOV words and disregarding them generally produces higher scores for \sm{Eng} joy \sm{and fear}. 

\noindent {\it Discussion:} Just as for valence, the method of handling OOVs becomes practically relevant only for smaller bin sizes, as for larger bin sizes the methods produce similar results \sm{overall}.

\subsection{Impact of Selectively Using the Lexicon}
\label{sec:thresh}

The continuously labeled emotion lexicons include words that may be very mildly associated with an emotion category or dimension. 
It is possible that the very low emotion association entries may in fact mislead the system, resulting in poor emotion arcs. 
We therefore investigate the quality of emotion arcs by generating them only from terms with an emotion association score greater than a pre-chosen threshold; thereby using the lexicon entries more selectively. 
We systematically vary the threshold to study what patterns of threshold lead to better arcs across the emotion test datasets.\footnote{Note that our goal is not to determine the predictive power of the system on new unseen test data. For that one would have to determine thresholds from a development set, and apply the model on unseen test data.}

Overall, we observed that valence benefits from including all terms, even lowly associated emotion words, as the optimal threshold across continuous and categorically datasets is 0 with a few notable exceptions (SemEval 2014 LiveJournal, SemEval 2014 tweets, and V-OC); Whereas anger, fear, joy and sadness, benefit from some degree of thresholding. For example, the optimal threshold for each is approximately: anger 0.75,
fear 0.25,  
joy 0.66,  
and 
sadness 0.66--0.75.

Generally, including only terms with emotion scores above 0.33 to 0.5 improves the quality of emotion arcs, with anger \sm{and sadness} preferring higher thresholds.
(Figures \ref{fig:v_cat_threshold_results} through \ref{fig:e_reg_threshold_results} in the Appendix show the full sets of correlations for reference) 
for valence and other emotion datasets. 

\begin{figure*}[t]
    \centering
    \includegraphics[width=0.75\textwidth]{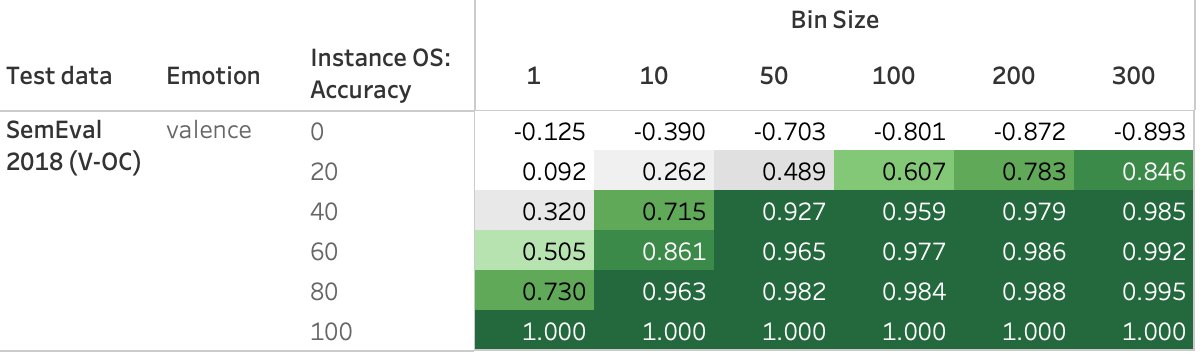}
    \caption{Valence: \sm{Spearman}
    correlations of the arcs generated using the Oracle System (OS) with the gold arcs created from the SemEval 2018 V-OC data.
    OS (Accuracy): Accuracy at instance-level sentiment classification.
    }
    \label{fig:oracle_results}
\end{figure*}

\begin{table*}[]
\centering
{\small
\begin{tabular}{llccccccc} 
\hline
\multicolumn{1}{l}{\textbf{Source}} &
\multicolumn{1}{l}{\textbf{Model}} &  \multicolumn{1}{c}{\textbf{Instance-Level}} & \multicolumn{6}{c}{\textbf{Bin Size}} \\
 & & \multicolumn{1}{c}{\textbf{Accuracy}} & 1 & 10 & 50 & 100 & 200 & 300 \\\hline
\citet{socher-etal-2013-recursive} & RNTN & 85.4\% & 0.829 & 0.972 & 0.980 & 0.983 & 0.986 & 0.992 \\ 
\citet{devlin-etal-2019-bert} & BERT-base & 93.5\% & 0.921 & 0.981 & 0.984 & 0.988 & 0.993 & 0.996\\ 
\citet{devlin-etal-2019-bert} & BERT-large & 94.9\% & 0.932 & 0.986 & 0.984 & 0.988 & 0.993 & 0.996 \\
\citet{xlnet} & XLNet & 97.0\% & 0.959 & 0.990 & 0.985 & 0.988 & 0.993 & 0.996\\
\citet{liu2020roberta} & RoBERTa & 96.4\% & 0.958 & 0.989 & 0.986 & 0.989 & 0.994 & 0.997\\ \hline

\end{tabular}
}
\caption{Valence: \sm{Spearman} 
correlations 
between arcs generated using various neural models for instance-level sentiment classification
and gold arcs created from the SemEval 2018 V-OC dataset.}
 \vspace*{-2mm}
\label{tab:flawed_oracle}
\end{table*}

\section{ML Arcs: Emotion Arcs Generated from Counting ML-Labeled Sentences}
\label{sec:flawed_oracle}

Even though the primary focus of this paper is to study how effective lexicon-based word-level emotion labeling methods are at generating emotion arcs, we devote this section to study how the accuracy of instance-level \sm{(sentence- or tweet-level)} emotion labeling impacts the quality of the generated emotion arcs.

We approached this 
by creating an `oracle' system, which has access to the gold instance emotion labels. \sm{There are several metrics for evaluating sentiment analysis at the instance level such as accuracy, correlation, or F-score. However we focus on accuracy as it is a simple intuitive metric.}
Inputs to the Oracle system are a dataset of text (for emotion labelling) and a level of accuracy \sm{such as 90\% accuracy (to perform instance-level emotion labelling at).}
Then, the system goes through each instance 
and predicts the correct emotion label with a probability corresponding to the pre-chosen accuracy.
In the case where the system decides to assign an incorrect label, it chooses one of the possible incorrect labels at random.
We use the oracle to generate emotion labels pertaining to various levels of accuracy, for the same dataset.
We then generate emotion arcs just 
as described in the previous section (by taking the average of the scores for each instance in a bin), and 
evaluate the generated arcs just as before (by determining correlation with the gold arc).  
This 
Oracle Systems allows us to see how 
accurate an instance level emotion labelling approaches needs to be to obtain various levels of quality when generating  
emotion arcs. 

Figure \ref{fig:oracle_results} shows the correlations of the valence arcs generated using the 
Oracle System with the gold valence arcs created from the 
SemEval 2018 V-OC
test set \sm{(that has 7 possible labels: -3 to 3)}.\footnote{Figures \ref{fig:appendix_oracle_cat_sem14} and \ref{fig:appendix_oracle_cat_sem18} (Appendix) show similar 
Oracle System results for \sm{other datasets}.}
We observe that, as expected the Oracle Systems with instance-level accuracy greater than approximately 
the random baseline (14.3\% for this dataset) 
obtain positive correlations;
whereas those with less than 14\% accuracy obtain negative correlations.
As seen with the results of the previous section, correlations increase markedly with increase in bin size. 
\sm{Even with an instance-level accuracy of 60\%}, correlation approaches 1
at 
larger 
bin sizes.  
Overall, we again observe  
high quality emotion arcs 
with bin sizes of just a few hundred instances.

Table \ref{tab:flawed_oracle} shows the correlations obtained on the same dataset when using 
various  
deep neural network and transformer-based systems. 
Observe that the recent systems obtain nearly perfect correlation at bin sizes 200 and 300.
That said, for applications where simple, interpretable, low-cost, and low-carbon-footprint systems are desired, the lexicon based systems described in the previous section, are perhaps more suitable.

\section{Conclusions}
This work, systematically and quantitatively, evaluated two broad 
approaches to automatically generating emotion arcs.
We showed that the arcs generated using lexicons,
and with bin sizes of just a few hundred instances, 
obtain very high correlations with the gold arcs.
The patterns remain consistent across a variety of datasets
and a number of emotion dimensions.
We also showed that the best neural ML models generate arcs that are close to perfect
with bin sizes of just a few hundred instances. 
However, the lexicon approach is competitive with those approaches,
and likely a better choice in applications that need high interpretability, low cost, low carbon foot print, and no need for domain-specific training.
This also allows a vast number of
researchers, from fields such as Psychology and Digital Humanities, who may not have the resources or necessary programming experience to deploy deep neural models, to still develop high-quality emotion arcs for their work. 

\section{Limitations}
This work has only explored estimating aggregate level emotion arcs 
by simply averaging the emotion scores of instances (in relevant bins / time steps). 
However, emotion arcs in coherent narratives, such as stories, are more complex, and may not be captured by this method. Understanding emotion arcs of narratives remains an open challenge.

\section{Ethics Considerations}
\noindent Our research interest is to study emotions at an aggregate/group level. This has applications in determining public policy (e.g., pandemic-response initiatives) and commerce (understanding attitudes towards products). However, emotions are complex, private, and central to an individual's experience. Additionally, each individual expresses emotion differently through language, which results in large amounts of variation. Therefore, several ethical considerations should be accounted for when performing any textual analysis of emotions \cite{Mohammad22AER,mohammad2020practical}.
The ones we would particularly like to highlight are listed below:
\begin{itemize}
\vspace*{-2mm}
    \item Our work on studying emotion word usage should not be construed as detecting how people feel; rather, we draw inferences on the emotions that are conveyed by users via the language that they use. 
\vspace*{-2mm}
    \item The language used in an utterance may convey information about the emotional state (or perceived emotional state) of the speaker, listener, or someone mentioned in the utterance. However, it is not sufficient for accurately determining any of their momentary emotional states. Deciphering true momentary emotional state of an individual requires extra-linguistic context and world knowledge.
    Even then, one can be easily mistaken.
\vspace*{-2mm}
    \item The inferences we draw in this paper are based on aggregate trends across large populations. We do not draw conclusions about specific individuals or momentary emotional states.
\vspace*{-2mm}
\end{itemize}

\bibliographystyle{acl_natbib}
\bibliography{anthology,custom}

\begin{thebibliography}{29}
\expandafter\ifx\csname natexlab\endcsname\relax\def\natexlab#1{#1}\fi

\bibitem[{Ashida et~al.(2021)Ashida, Tokumaru, and Kojiri}]{writenovels}
Atsushi Ashida, Masataka Tokumaru, and Tomoko Kojiri. 2021.
\newblock \href {https://doi.org/10.12937/itel.1.1.Reg.p005} {Characters’
  emotion design support system for writing novels based on story arcs of
  target readers}.
\newblock \emph{Information and Technology in Education and Learning},
  1:Reg--p005.

\bibitem[{Bhyravajjula et~al.(2022)Bhyravajjula, Narayan, and
  Shrivastava}]{Bhyravajjula2022MARCUSAE}
Sriharsh Bhyravajjula, Ujwal Narayan, and Manish Shrivastava. 2022.
\newblock Marcus: An event-centric nlp pipeline that generates character arcs
  from narratives.
\newblock In \emph{Text2Story@ECIR}.

\bibitem[{Brahman and Chaturvedi(2020)}]{brahman-chaturvedi-2020-modeling}
Faeze Brahman and Snigdha Chaturvedi. 2020.
\newblock \href {https://doi.org/10.18653/v1/2020.emnlp-main.426} {Modeling
  protagonist emotions for emotion-aware storytelling}.
\newblock In \emph{Proceedings of the 2020 Conference on Empirical Methods in
  Natural Language Processing (EMNLP)}, pages 5277--5294, Online. Association
  for Computational Linguistics.

\bibitem[{Chong and Gottipati(2020)}]{socialmedia}
Mark Chong and Swapna Gottipati. 2020.
\newblock \href
  {https://login.ezproxy.library.ualberta.ca/login?url=https://www.proquest.com/scholarly-journals/social-media-influences-lnstagram-storytelling/docview/2496342719/se-2?accountid=14474}
  {Social media influences and instagram storytelling: Case study of singapore
  instagram influences}.
\newblock \emph{The Journal of Applied Business and Economics}, 22(10):81--96.
\newblock Copyright - Copyright North American Business Press 2020; Last
  updated - 2021-06-03.

\bibitem[{Devlin et~al.(2019)Devlin, Chang, Lee, and
  Toutanova}]{devlin-etal-2019-bert}
Jacob Devlin, Ming-Wei Chang, Kenton Lee, and Kristina Toutanova. 2019.
\newblock \href {https://doi.org/10.18653/v1/N19-1423} {{BERT}: Pre-training of
  deep bidirectional transformers for language understanding}.
\newblock In \emph{Proceedings of the 2019 Conference of the North {A}merican
  Chapter of the Association for Computational Linguistics: Human Language
  Technologies, Volume 1 (Long and Short Papers)}, pages 4171--4186,
  Minneapolis, Minnesota. Association for Computational Linguistics.

\bibitem[{Ding and Riloff(2018)}]{10.5555/3504035.3504742}
Haibo Ding and Ellen Riloff. 2018.
\newblock Weakly supervised induction of affective events by optimizing
  semantic consistency.
\newblock In \emph{Proceedings of the Thirty-Second AAAI Conference on
  Artificial Intelligence and Thirtieth Innovative Applications of Artificial
  Intelligence Conference and Eighth AAAI Symposium on Educational Advances in
  Artificial Intelligence}, AAAI'18/IAAI'18/EAAI'18. AAAI Press.

\bibitem[{Feng et~al.(2013)Feng, Kang, Kuznetsova, and
  Choi}]{feng-etal-2013-connotation}
Song Feng, Jun~Seok Kang, Polina Kuznetsova, and Yejin Choi. 2013.
\newblock \href {https://aclanthology.org/P13-1174} {Connotation lexicon: A
  dash of sentiment beneath the surface meaning}.
\newblock In \emph{Proceedings of the 51st Annual Meeting of the Association
  for Computational Linguistics (Volume 1: Long Papers)}, pages 1774--1784,
  Sofia, Bulgaria. Association for Computational Linguistics.

\bibitem[{Hipson and Mohammad(2021)}]{moviedialogues}
Will~E. Hipson and Saif~M. Mohammad. 2021.
\newblock \href {https://doi.org/10.1371/journal.pone.0256153} {Emotion
  dynamics in movie dialogues}.
\newblock \emph{PLOS ONE}, 16(9):1--19.

\bibitem[{Kim et~al.(2017)Kim, Pad{\'o}, and
  Klinger}]{kim-etal-2017-investigating}
Evgeny Kim, Sebastian Pad{\'o}, and Roman Klinger. 2017.
\newblock \href {https://doi.org/10.18653/v1/W17-2203} {Investigating the
  relationship between literary genres and emotional plot development}.
\newblock In \emph{Proceedings of the Joint {SIGHUM} Workshop on Computational
  Linguistics for Cultural Heritage, Social Sciences, Humanities and
  Literature}, pages 17--26, Vancouver, Canada. Association for Computational
  Linguistics.

\bibitem[{Liu et~al.(2020)Liu, Ott, Goyal, Du, Joshi, Chen, Levy, Lewis,
  Zettlemoyer, and Stoyanov}]{liu2020roberta}
Yinhan Liu, Myle Ott, Naman Goyal, Jingfei Du, Mandar Joshi, Danqi Chen, Omer
  Levy, Mike Lewis, Luke Zettlemoyer, and Veselin Stoyanov. 2020.
\newblock \href {https://arxiv.org/abs/1907.11692} {Ro{\{}bert{\}}a: A robustly
  optimized {\{}bert{\}} pretraining approach}.

\bibitem[{Mohammad(2011)}]{mohammad-2011-upon}
Saif Mohammad. 2011.
\newblock \href {https://aclanthology.org/W11-1514} {From once upon a time to
  happily ever after: Tracking emotions in novels and fairy tales}.
\newblock In \emph{Proceedings of the 5th {ACL}-{HLT} Workshop on Language
  Technology for Cultural Heritage, Social Sciences, and Humanities}, pages
  105--114, Portland, OR, USA. Association for Computational Linguistics.

\bibitem[{Mohammad(2018{\natexlab{a}})}]{vad-acl2018}
Saif~M. Mohammad. 2018{\natexlab{a}}.
\newblock Obtaining reliable human ratings of valence, arousal, and dominance
  for 20,000 english words.
\newblock In \emph{Proceedings of The Annual Conference of the Association for
  Computational Linguistics (ACL)}, Melbourne, Australia.

\bibitem[{Mohammad(2018{\natexlab{b}})}]{LREC18-AIL}
Saif~M. Mohammad. 2018{\natexlab{b}}.
\newblock Word affect intensities.
\newblock In \emph{Proceedings of the 11th Edition of the Language Resources
  and Evaluation Conference (LREC-2018)}, Miyazaki, Japan.

\bibitem[{Mohammad(2020)}]{mohammad2020practical}
Saif~M. Mohammad. 2020.
\newblock \href {http://arxiv.org/abs/2011.03492} {Practical and ethical
  considerations in the effective use of emotion and sentiment lexicons}.

\bibitem[{Mohammad(2022)}]{Mohammad22AER}
Saif~M. Mohammad. 2022.
\newblock Ethics sheet for automatic emotion recognition and sentiment
  analysis.
\newblock \emph{To Appear in Computational Linguistics}.

\bibitem[{Mohammad et~al.(2018)Mohammad, Bravo-Marquez, Salameh, and
  Kiritchenko}]{SemEval2018Task1}
Saif~M. Mohammad, Felipe Bravo-Marquez, Mohammad Salameh, and Svetlana
  Kiritchenko. 2018.
\newblock Semeval-2018 {T}ask 1: {A}ffect in tweets.
\newblock In \emph{Proceedings of International Workshop on Semantic Evaluation
  (SemEval-2018)}, New Orleans, LA, USA.

\bibitem[{Mohammad and Turney(2013)}]{Mohammad13}
Saif~M. Mohammad and Peter~D. Turney. 2013.
\newblock Crowdsourcing a word-emotion association lexicon.
\newblock \emph{Computational Intelligence}, 29(3):436--465.

\bibitem[{Osgood et~al.(1957)Osgood, Suci, and Tannenbaum}]{Osgood57}
Charles~E Osgood, George~J Suci, and Percy Tannenbaum. 1957.
\newblock \emph{The measurement of meaning}.
\newblock University of Illinois Press.

\bibitem[{Park and Kwak(2020)}]{cellphone}
Geunseok Park and Minjung Kwak. 2020.
\newblock \href {https://doi.org/10.24507/icicelb.11.05.509} {The life cycle of
  online smartphone reviews: Investigating dynamic change in customer opinion
  using sentiment analysis}.
\newblock \emph{ICIC Express Letters}.

\bibitem[{Reagan et~al.(2016)Reagan, Mitchell, Kiley, Danforth, and
  Dodds}]{emotionarcs}
Andrew~J. Reagan, Lewis Mitchell, Dilan Kiley, Christopher~M. Danforth, and
  Peter~S. Dodds. 2016.
\newblock \href
  {https://www.proquest.com/scholarly-journals/emotional-arcs-stories-are-dominated-six-basic/docview/1865288690/se-2}
  {The emotional arcs of stories are dominated by six basic shapes}.
\newblock \emph{EPJ Data Science}, 5(1):1--12.
\newblock Copyright - EPJ Data Science is a copyright of Springer, 2016; Last
  updated - 2017-02-06.

\bibitem[{Rosenthal et~al.(2014)Rosenthal, Ritter, Nakov, and
  Stoyanov}]{rosenthal-etal-2014-semeval}
Sara Rosenthal, Alan Ritter, Preslav Nakov, and Veselin Stoyanov. 2014.
\newblock \href {https://doi.org/10.3115/v1/S14-2009} {{S}em{E}val-2014 task 9:
  Sentiment analysis in {T}witter}.
\newblock In \emph{Proceedings of the 8th International Workshop on Semantic
  Evaluation ({S}em{E}val 2014)}, pages 73--80, Dublin, Ireland. Association
  for Computational Linguistics.

\bibitem[{Russell(2003)}]{russell2003core}
James~A Russell. 2003.
\newblock Core affect and the psychological construction of emotion.
\newblock \emph{Psychological review}, 110(1):145.

\bibitem[{Russell and Mehrabian(1977)}]{russell1977evidence}
James~A Russell and Albert Mehrabian. 1977.
\newblock Evidence for a three-factor theory of emotions.
\newblock \emph{Journal of research in Personality}, 11(3):273--294.

\bibitem[{Seidl(2020)}]{Seidl2020ThePO}
Timo Seidl. 2020.
\newblock \href {https://doi.org/https://doi.org/10.1111/rego.12353} {The
  politics of platform capitalism. a case study on the regulation of uber in
  new york}.
\newblock \emph{SocArXiv}.

\bibitem[{Singh et~al.(2020)Singh, Dwivedi, Kahlon, Sawhney, Alalwan, and
  Rana}]{policy}
Prabhsimran Singh, Yogesh Dwivedi, Karanjeet Kahlon, Dr. Ravinder~Singh
  Sawhney, Ali Alalwan, and Nripendra Rana. 2020.
\newblock \href {https://doi.org/10.1007/s10796-019-09916-y} {Smart monitoring
  and controlling of government policies using social media and cloud
  computing}.
\newblock \emph{Information Systems Frontiers}, 22.

\bibitem[{Socher et~al.(2013)Socher, Perelygin, Wu, Chuang, Manning, Ng, and
  Potts}]{socher-etal-2013-recursive}
Richard Socher, Alex Perelygin, Jean Wu, Jason Chuang, Christopher~D. Manning,
  Andrew Ng, and Christopher Potts. 2013.
\newblock \href {https://aclanthology.org/D13-1170} {Recursive deep models for
  semantic compositionality over a sentiment treebank}.
\newblock In \emph{Proceedings of the 2013 Conference on Empirical Methods in
  Natural Language Processing}, pages 1631--1642, Seattle, Washington, USA.
  Association for Computational Linguistics.

\bibitem[{Somasundaran et~al.(2020)Somasundaran, Chen, and
  Flor}]{somasundaran-etal-2020-emotion}
Swapna Somasundaran, Xianyang Chen, and Michael Flor. 2020.
\newblock \href {https://doi.org/10.18653/v1/2020.nuse-1.12} {Emotion arcs of
  student narratives}.
\newblock In \emph{Proceedings of the First Joint Workshop on Narrative
  Understanding, Storylines, and Events}, pages 97--107, Online. Association
  for Computational Linguistics.

\bibitem[{Spearman(1987)}]{spearman1987proof}
Charles Spearman. 1987.
\newblock The proof and measurement of association between two things.
\newblock \emph{The American journal of psychology}, 100(3/4):441--471.

\bibitem[{Yang et~al.(2019)Yang, Dai, Yang, Carbonell, Salakhutdinov, and
  Le}]{xlnet}
Zhilin Yang, Zihang Dai, Yiming Yang, Jaime Carbonell, Russ~R Salakhutdinov,
  and Quoc~V Le. 2019.
\newblock \href
  {https://proceedings.neurips.cc/paper/2019/file/dc6a7e655d7e5840e66733e9ee67cc69-Paper.pdf}
  {Xlnet: Generalized autoregressive pretraining for language understanding}.
\newblock In \emph{Advances in Neural Information Processing Systems},
  volume~32. Curran Associates, Inc.

\end{thebibliography}

\appendix
\noindent \textbf{APPENDIX}

\noindent \sm{To determine the generality of the claims made in the paper, we conducted experiments on 36
emotion labeled datasets from three languages with labels that are categorical and continuous for five affect categories.
While the core representative experiments are presented in the main paper, remaining results tables are available in the Appendix below. The results on individual datasets also establish key benchmarks that be useful for 
researchers and developers to assess the expected quality of the emotion arcs under various settings. Readers can examine individual tables of interest to them. 

Appendix \ref{appendix:data_descriptive} 
presents additional dataset summary statistics. 
The next four appendices present correlation scores between the gold arcs and the generated arcs:
first with the full lexicons and the continuous datasets (Appendix \ref{appendix:v_cont} and \ref{appendix:e_cont});
and then with the thresholded lexicons and both the categorical and continuous datasets 
(Appendix \ref{appendix:val_threshold} and \ref{appendix:four_e_threshold}).
Results using the full lexicon and the categorical datasets are in the main paper.
}

\section{Data Descriptive Statistics}
\label{appendix:data_descriptive}
\sm{
Tables \ref{tab:datasets_arabic} and \ref{tab:datasets_spanish} show key summarizing information about the Arabic and Spanish datasets, respectively. Table \ref{tab:app_datasets} shows the average number of words per instance in each of the Arabic, English, and Spanish datasets.}

\begin{table*}[t]
\centering
{\small
\begin{tabular}
{lllllll}
\hline
\multicolumn{1}{l}{\textbf{Dataset}} &
\multicolumn{1}{l}{\textbf{Source}} &
\multicolumn{1}{l}{\textbf{Domain}} &
\multicolumn{1}{l}{\textbf{Dimension}} &
\multicolumn{1}{l}{\textbf{Label Type}} & \multicolumn{1}{l}{\textbf{\# Instances}}\\ \hline
SemEval 2018 (EI-OC) & \citet{SemEval2018Task1} & tweets & anger, fear,  & categorical (0, 1, 2, 3) & 1400 \\
    &  &   & joy, sadness   &  & \\[3pt] 

SemEval 2018 (EI-Reg) & \citet{SemEval2018Task1} & tweets & anger, fear & continuous (0 to 1) & 1400\\
     &  &   & joy, sadness   &  &  \\[3pt]

SemEval 2018 (V-OC) &  \citet{SemEval2018Task1} & tweets & valence & categorical (-3,-2,...3) & 1800\\
SemEval 2018 (V-Reg) &  \citet{SemEval2018Task1} & tweets & valence & continuous (0 to 1) & 1800\\\hline

\end{tabular}
}
\caption{Dataset descriptive statistics for datasets in Arabic. The No. of instances includes the train, development, and test sets for the Sem-Eval 2018 Task 1 (EI-OC, EI-Reg, V-OC and V-Reg).} 
 \label{tab:datasets_arabic}
\end{table*}

\begin{table*}[t]
\centering
{\small
\begin{tabular}
{lllllll}
\hline
\multicolumn{1}{l}{\textbf{Dataset}} &
\multicolumn{1}{l}{\textbf{Source}} &
\multicolumn{1}{l}{\textbf{Domain}} &
\multicolumn{1}{l}{\textbf{Dimension}} &
\multicolumn{1}{l}{\textbf{Label Type}} & \multicolumn{1}{l}{\textbf{\# Instances}}\\ \hline
SemEval 2018 (EI-OC) & \citet{SemEval2018Task1} & tweets & anger, fear,  & categorical (0, 1, 2, 3) & 1986, 1986\\
    &  &   & joy, sadness &  & 1990, 1991\\[3pt]

SemEval 2018 (EI-Reg) & \citet{SemEval2018Task1} & tweets & anger, fear & continuous (0 to 1) & 1986, 1986\\
     &  &   & joy, sadness   &  & 1990, 1991\\[3pt]

SemEval 2018 (V-OC) &  \citet{SemEval2018Task1} & tweets & valence & categorical (-3,-2,...3) & 2443\\
SemEval 2018 (V-Reg) &  \citet{SemEval2018Task1} & tweets & valence & continuous (0 to 1) & 2443\\\hline

\end{tabular}
}
\caption{Dataset descriptive statistics for datasets in Spanish. The No. of instances includes the train, development, and test sets for the Sem-Eval 2018 Task 1 (EI-OC, EI-Reg, V-OC and V-Reg).} 
 \label{tab:datasets_spanish}
\end{table*}

\begin{table*}[t!]
\centering
{\small
\begin{tabular}{lllll}
\hline
\multicolumn{1}{l}{\textbf{Dataset}} &
\multicolumn{1}{l}{\textbf{Lang}} &
\multicolumn{1}{l}{\textbf{Source}} &
\multicolumn{1}{l}{\textbf{Domain}} &
\multicolumn{1}{l}{\textbf{Avg. \# Words per Instance}}\\ \hline
Movie Reviews Categorical/Continuous & Eng & \citet{socher-etal-2013-recursive} & movie & 16.38 \\

SemEval 2014 & Eng & \citet{rosenthal-etal-2014-semeval} & Multiple$^*$ & 16.07\\[3pt]

SemEval 2018 (EI-OC/EI-Reg) & Eng & \citet{SemEval2018Task1} & tweets & 13.29\\[3pt]

SemEval 2018 (V-OC/V-Reg) & Eng & \citet{SemEval2018Task1} & tweets & 13.05\\ [3pt]

SemEval 2018 (EI-OC/EI-Reg) & Ar & \citet{SemEval2018Task1} & tweets & 13.29\\[3pt]

SemEval 2018 (V-OC/V-Reg) & Ar &  \citet{SemEval2018Task1} & tweets & 13.05\\[3pt]

SemEval 2018 (EI-OC/EI-Reg) & Es & \citet{SemEval2018Task1} & tweets & 13.29\\[3pt]

SemEval 2018 (V-OC/V-Reg) & Es &  \citet{SemEval2018Task1} & tweets & 13.05\\\hline

\end{tabular}
}
\caption{  
The average number of words per instance includes the train, development, and test sets for the Sem-Eval 2014 Task 9 and the Sem-Eval 2018 Task 1 (EI-OC, EI-Reg, V-OC and V-Reg), and considers the data after tokenizing and only words composed of alphabet letters. 
Multiple$^*$: The SemEval 2014 dataset has collections of LiveJournal posts (1141), SMS messages (2082), regular tweets (15302), and sarcastic tweets (86) for a total of 18611 instances. The Movie Reviews Categorical dataset is often also referred to as SST-2.} 
 \label{tab:app_datasets}
\end{table*}


\section{Gold--Auto Correlations for Valence and Continuously Labeled Test Data}
\label{appendix:v_cont}
\begin{figure*}[t!]
    \centering
    \includegraphics[width=0.85\textwidth]{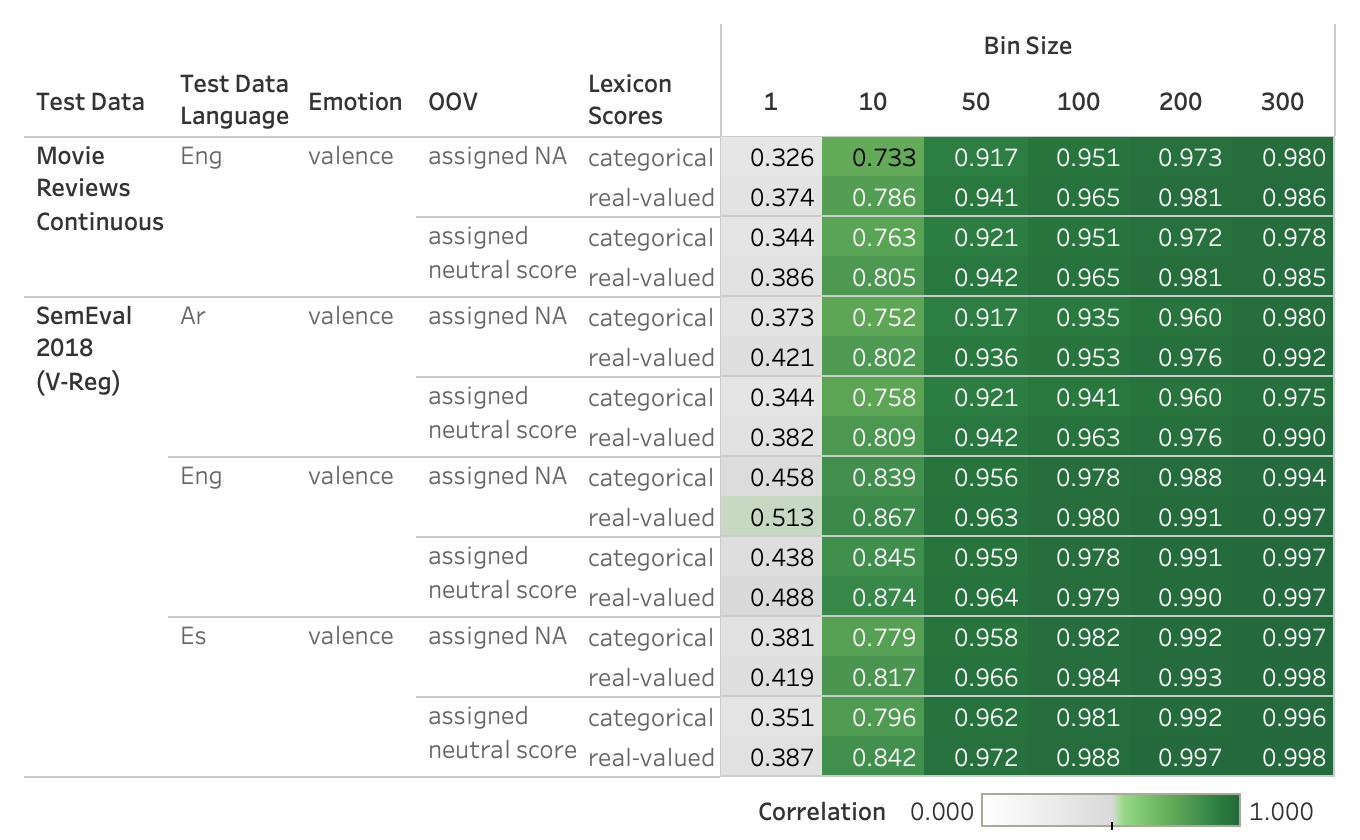}
    \caption{Valence: 
    \sm{Spearman} correlations between 
    arcs generated using lexicons and 
    gold arcs
    created from \textit{continuously} labeled  test data.}
    \label{fig:v_reg_results}
    \vspace*{-4mm}
\end{figure*}

Figure \ref{fig:v_reg_results} shows the Spearman correlations between automatically generated emotion arcs and arcs created from human annotations for valence from \textit{continuously} labeled test data.
It shows similar patterns as those discussed in the paper for categorical data overall.

\section{Gold--Auto Correlations for Anger, Fear, Joy, Sadness and Continuously Labeled Test Data}
\label{appendix:e_cont}
\begin{figure*}[t]
    \centering
    \includegraphics[width=0.9\textwidth]{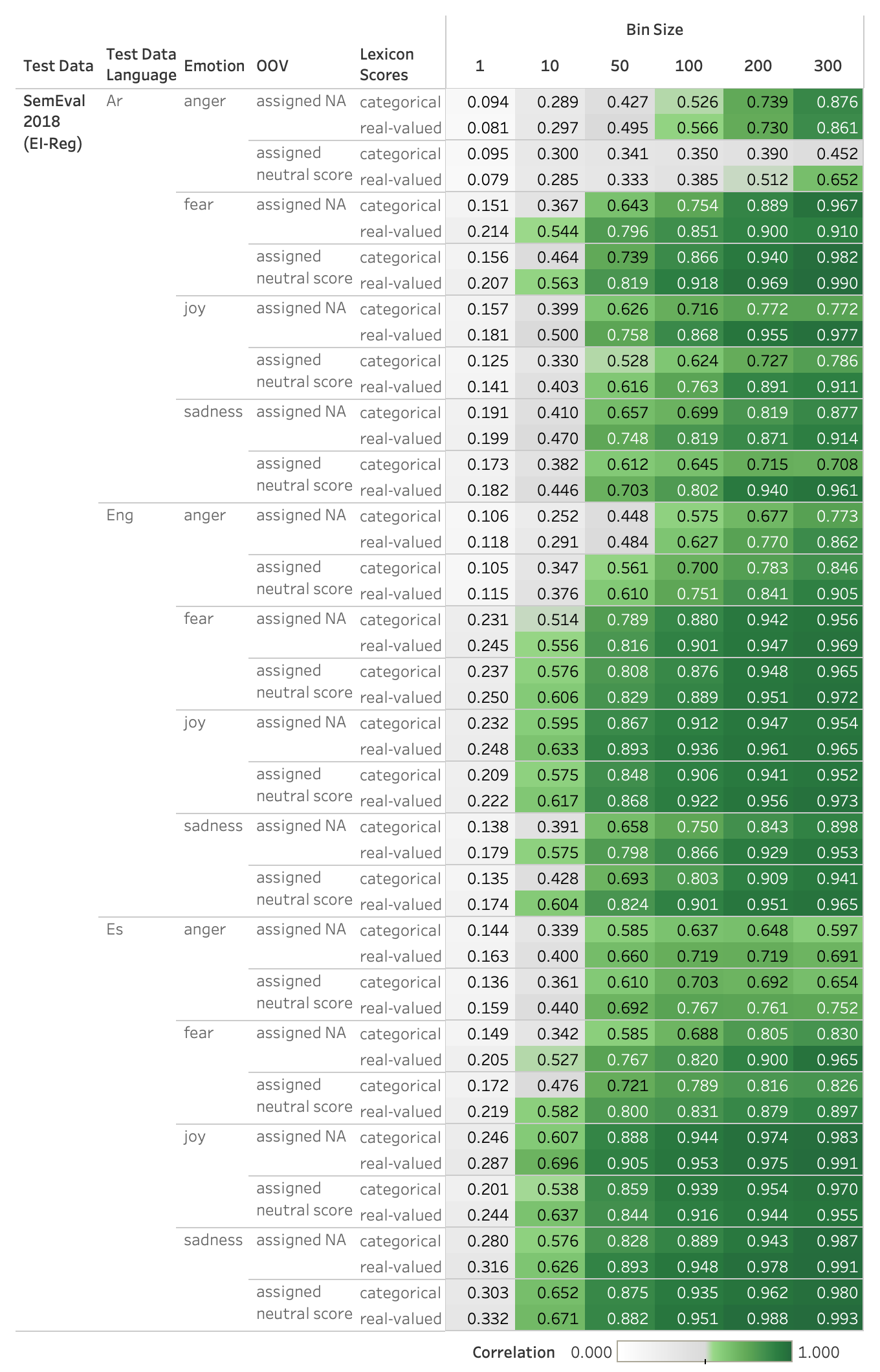}
     \caption{Anger, fear, joy and sadness: \sm{Spearman} correlations between arcs generated using lexicons and 
    gold arcs
    created from \textit{continuously} labeled test data.}
    
       \vspace*{-6mm}
    \label{fig:e_reg_results}

\end{figure*}

Figure \ref{fig:e_reg_results} shows the Spearman correlations between automatically generated emotion arcs and arcs created from human annotations for anger, fear, joy and sadness arcs from \textit{continuously} labeled test data.
It shows similar patterns as those discussed in the paper for categorical data overall.

\section{Gold--Auto Correlations for Valence and Thresholded Lexicons}
\label{appendix:val_threshold}
\begin{figure*}[t]
    \centering
    \includegraphics[width=0.9\textwidth]{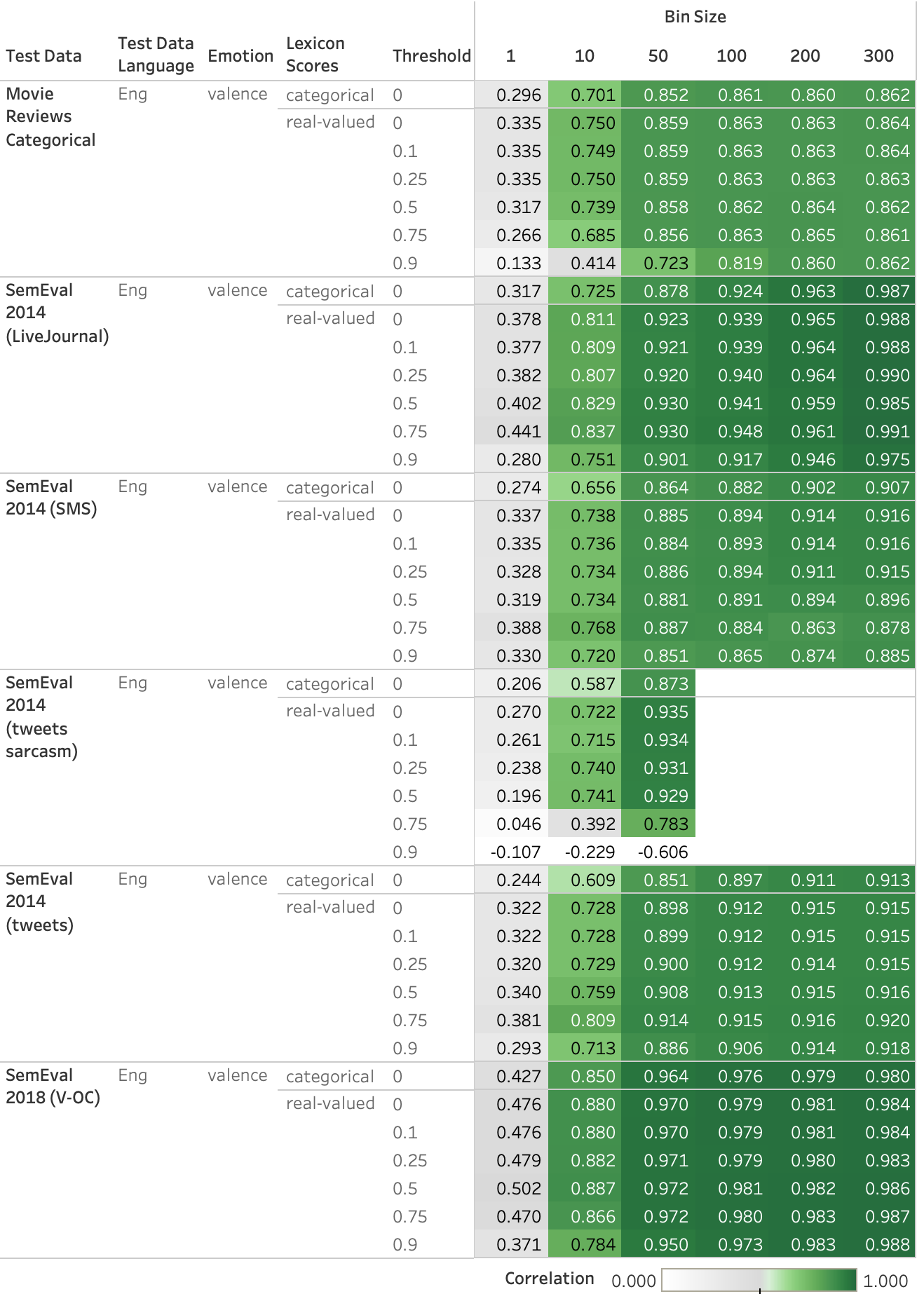}
    \caption{Valence: 
    \sm{Spearman} correlations between arcs generated using lexicons only containing terms with an emotion score above the given \textbf{threshold} and gold arcs created from \textit{categorically} labeled test data.}
    \label{fig:v_cat_threshold_results}
\end{figure*}

Figure \ref{fig:v_cat_threshold_results} shows the 
correlations between generated emotion arcs and arcs created using human annotations when only considering terms which have an emotion score in the lexicon above a given threshold. We first look at the datasets for valence which have categorical and then continuous labels.

The optimal threshold for the highest correlation results 
across all categorically labeled datasets is close to 0 with a few notable exceptions (SemEval 2014 LiveJournal, SemEval 2014 tweets, and V-OC). This means that when generating emotion arcs for valence, it is beneficial to consider mostly all terms regardless of their emotion scores as this information builds higher quality emotion arcs.

\begin{figure*}[t]
    \centering
    \includegraphics[width=0.85\textwidth]{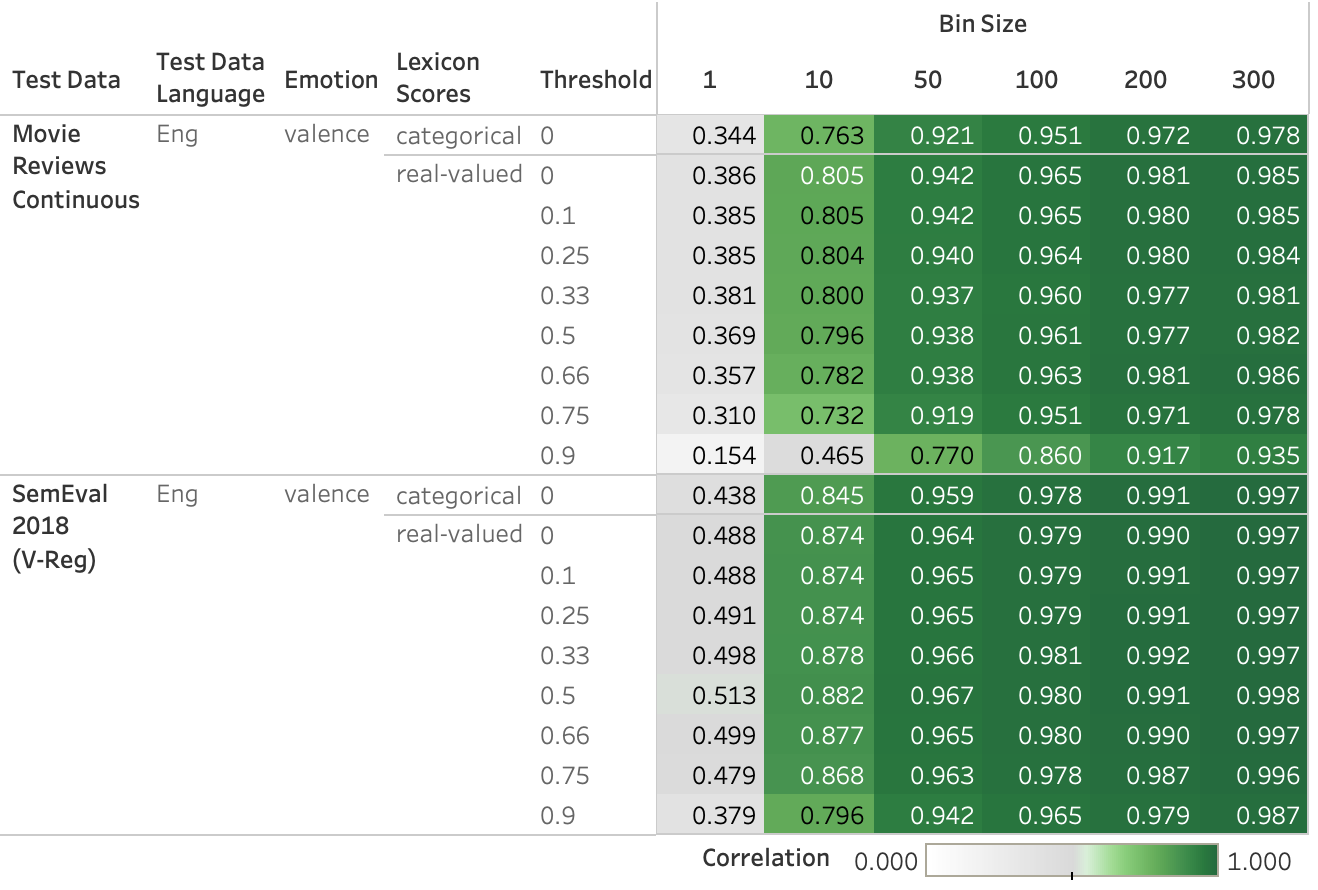}
    \caption{Valence: 
    \sm{Spearman}
    correlations between arcs generated using lexicons only containing terms with an emotion score above the given \textbf{threshold} and gold arcs created from \textit{continuously} labeled test data.}
    \label{fig:v_reg_threshold_results}
\end{figure*}

 Likewise, a similar pattern holds for valence with continuous labels as shown in Figure \ref{fig:v_reg_threshold_results}. The optimal thresholds for these datasets is 0 to 0.33-0.50. 

\section{Gold--Auto Correlations for Anger, Fear, Joy, Sadness and Thresholded Lexicons}
\label{appendix:four_e_threshold}
\begin{figure*}[t]
    \centering
    \includegraphics[width=0.9\textwidth]{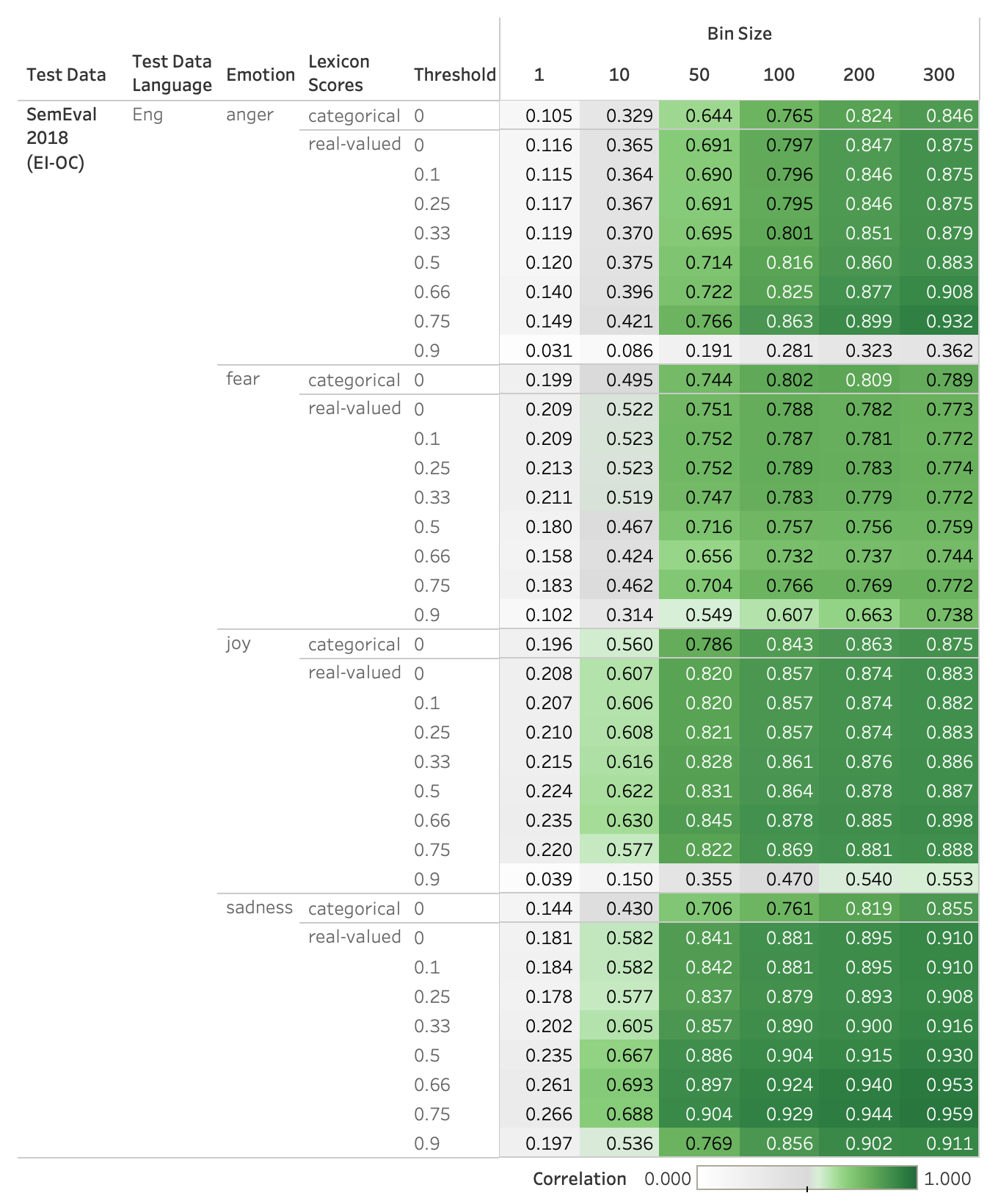}
    \caption{Anger, fear, joy and sadness: 
    \sm{Spearman}
    correlations between arcs generated using lexicons only containing terms with an emotion score above the given \textbf{threshold} and gold arcs created from \textit{categorically} labeled test data.}
    \label{fig:e_cat_threshold_results}
\end{figure*}

Figure \ref{fig:e_cat_threshold_results} shows the correlations between generated emotion arcs and those created using human annotations when only considering terms which have an emotion score in the lexicon above a given threshold for anger, fear, joy and sadness in categorical labeled datasets.

The optimal threshold producing the highest correlations differs based on emotion. For anger it is 0.75, for fear 0.25, 
for joy 0.66, and lastly for sadness 0.75. 
These results contrast those found with valence, where the optimal threshold was close to 0. When generating emotion arcs for anger, fear, joy, and sadness they benefit from including only more highly associated emotion terms.

\begin{figure*}[t]
    \centering
    \includegraphics[width=0.9\textwidth]{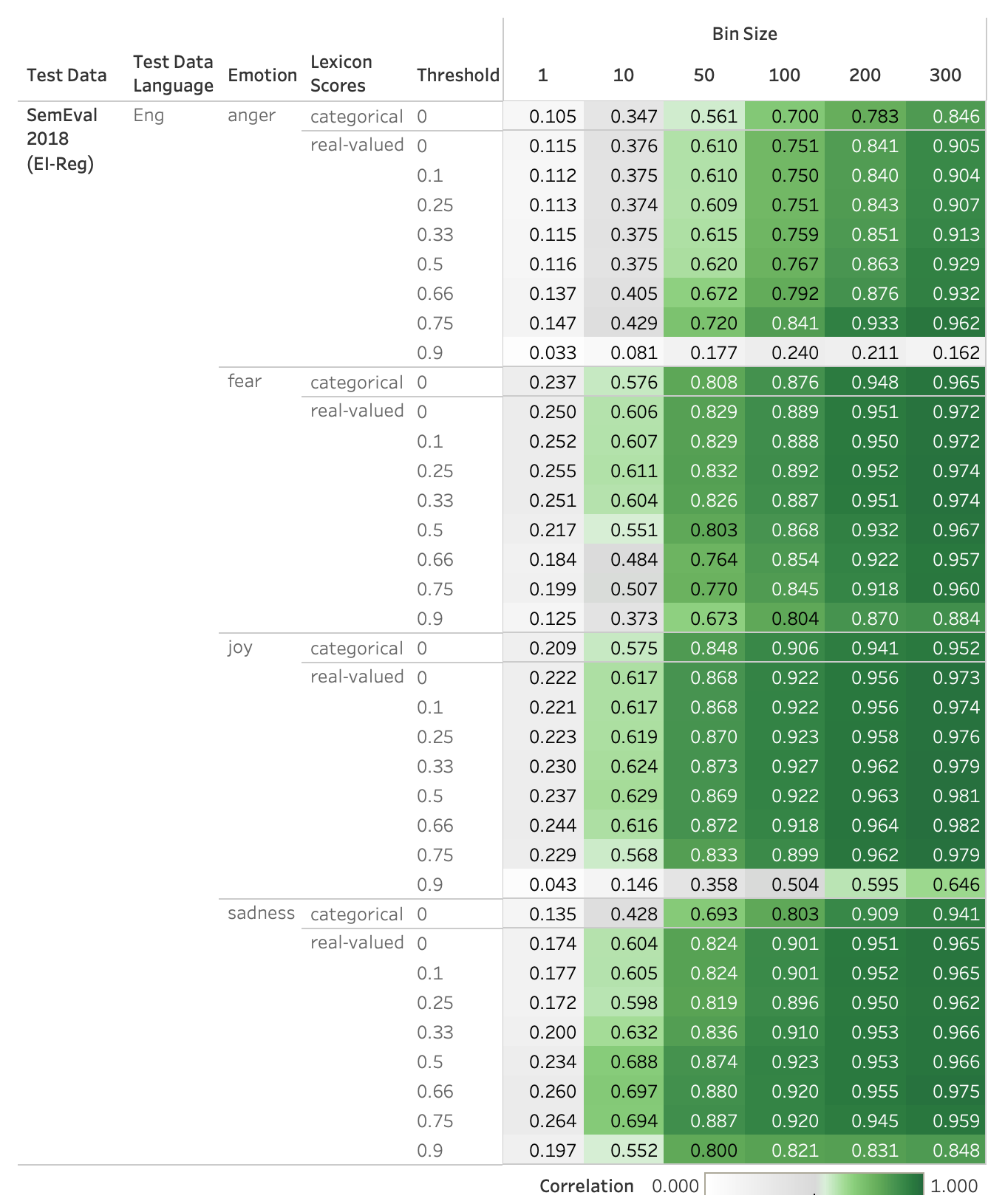}
    \caption{Anger, fear, joy and sadness: 
    \sm{Spearman}
    correlations between arcs generated using lexicons only containing terms with an emotion score above the given \textbf{threshold} and gold arcs created from \textit{continuously} labeled test data.}
    \label{fig:e_reg_threshold_results}
\end{figure*}

In Figure \ref{fig:e_reg_threshold_results} we show the
results
for anger, fear, joy and sadness with continuously labeled datasets. 
Overall, these optimal \textbf{thresholds} are similar to those for categorical labeled dataset. 

\section{Impact of the Quality of Instance-Level Emotion Labeling on Emotion Arcs}
\label{sec:appendix_oracle_cat}

\begin{figure*}[t]
    \centering
    \includegraphics[width=0.85\textwidth]{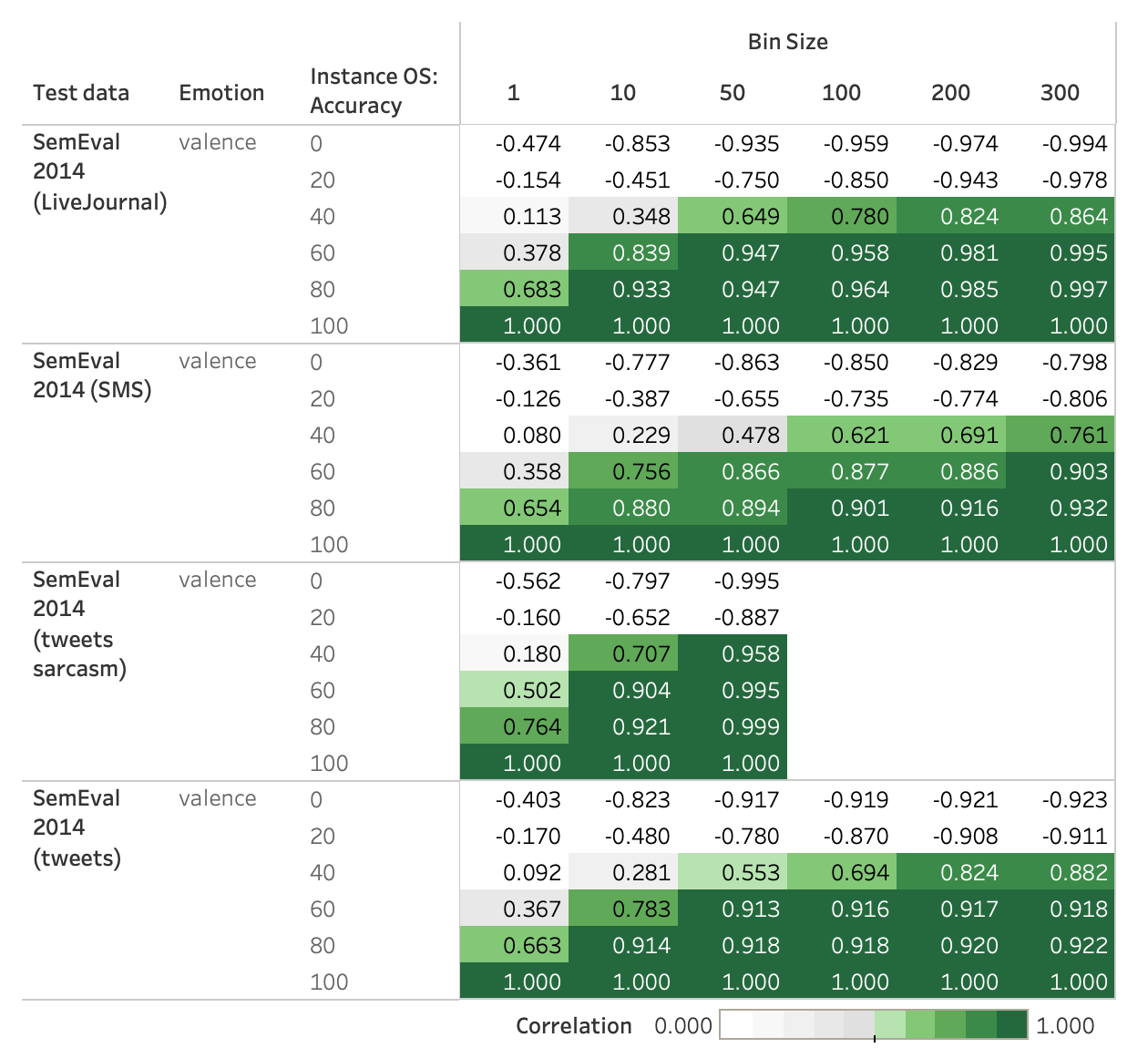}
    \caption{
    Valence: \sm{Spearman}
    correlations of the arcs generated using the Oracle System with the gold arcs created from the \textit{categorically} labeled Sem-Eval 2014 test data. Note: ‘OS (Accuracy)’ refers to the accuracy of the Oracle System on instance-level sentiment classification.
    }
    \label{fig:appendix_oracle_cat_sem14}
\end{figure*}

\begin{figure*}[t]
    \centering
    \includegraphics[width=0.85\textwidth]{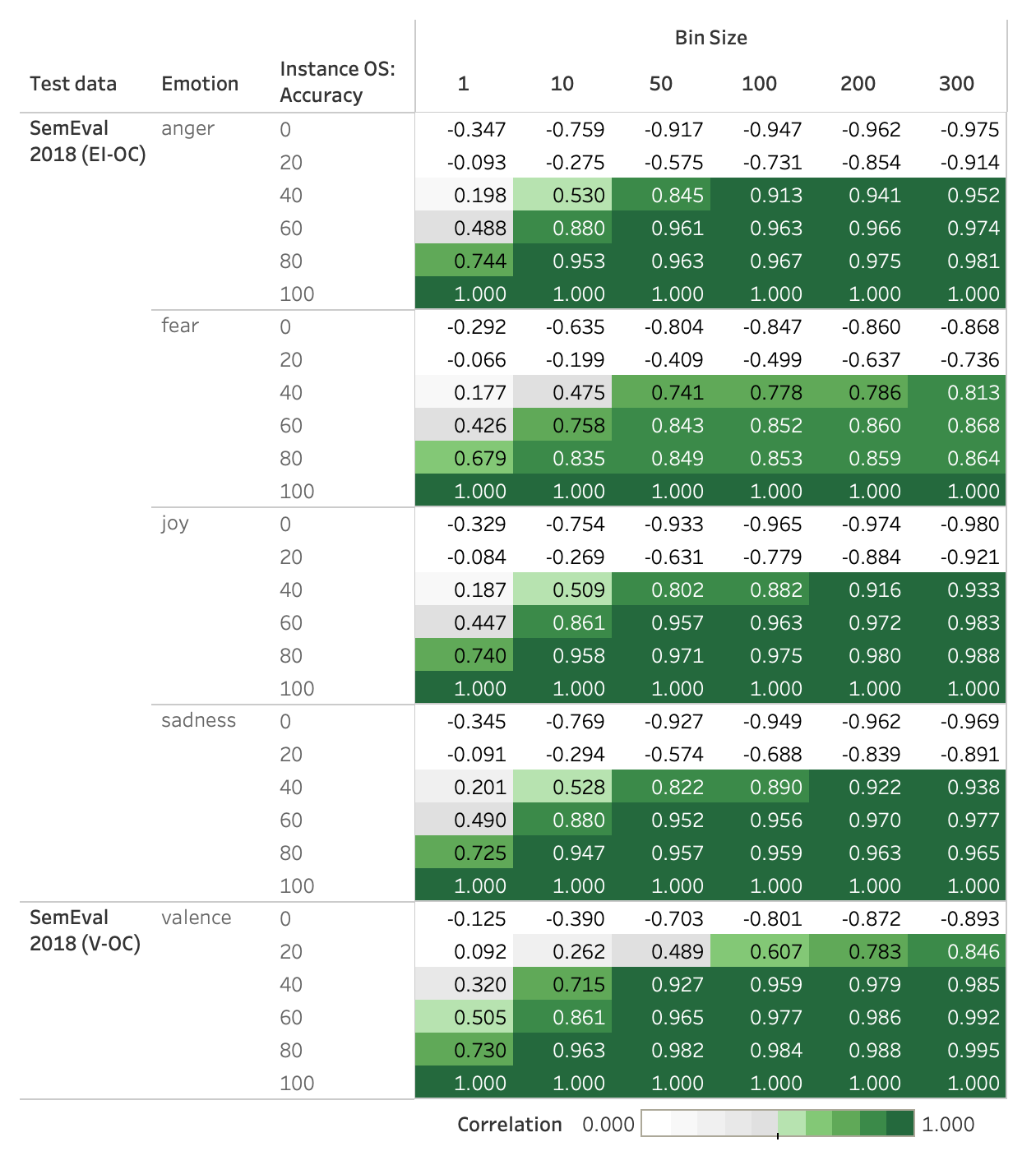}
    \caption{
    Emotions: \sm{Spearman} 
    correlations of the arcs generated using the Oracle System with the gold arcs created from the \textit{categorically} labeled Sem-Eval 2018 test data. Note: ‘OS (Accuracy)’ refers to the accuracy of the Oracle System on instance-level sentiment classification.
    }
    
    \label{fig:appendix_oracle_cat_sem18}
\end{figure*}

Figures \ref{fig:appendix_oracle_cat_sem14} and \ref{fig:appendix_oracle_cat_sem18} show the results for the Oracle System on the categorically labeled SemEval 2014 and SemEval 2018 datasets.
We observe similar patterns as discussed in the paper for the Movie Reviews dataset.
(Note that the instance-level random-guess baseline is dependent on the number of class labels;
thus, the minimum Oracle System Accuracy at which positive correlations with gold arcs appear is different across the datasets.) 

\end{document}